 \documentclass{article}
 \usepackage{icml2001}
 \usepackage{epsfig}

\begin{document} 

\twocolumn[
\icmltitle{Efficient algorithms for decision tree cross-validation}
 
\icmlauthor{Hendrik Blockeel}{Hendrik.Blockeel@cs.kuleuven.ac.be}
\icmlauthor{Jan Struyf}{Jan.Struyf@cs.kuleuven.ac.be}
\icmladdress{Katholieke Universiteit Leuven, Dept. of Computer Science,
Celestijnenlaan 200A, B-3001 Leuven, Belgium}
\vskip 0.3in
]

\begin{abstract} 
Cross-validation is a useful and generally applicable technique
often employed in machine learning, including decision
tree induction.  An important disadvantage of straightforward implementation
of the technique is its computational overhead.  In this paper 
we show that, for decision trees, the computational overhead of 
cross-validation can be reduced significantly by integrating the
cross-validation with the normal decision tree induction process.
We discuss how existing decision tree algorithms can be adapted to this
aim, and provide an analysis of the speedups these adaptations
may yield.  The analysis is supported by experimental results.
\end{abstract} 

\section{Introduction}

Cross-validation is a generally applicable and very useful technique for 
many tasks often encountered in machine learning, such as accuracy
estimation, feature selection or parameter tuning.  
It consists of partitioning a data set $D$ into $n$ subsets 
$D_i$ and then running a given algorithm $n$ times, each time using a 
different training set $D - D_i$ and validating the results on $D_i$.

Cross-validation is used within a wide range of machine learning
approaches, such as instance based learning, artificial neural
networks, or decision tree induction.  As an example of its use within
decision tree induction, the CART system \cite{Breiman84:other}
employs a tree pruning method that is based on trading off predictive
accuracy versus tree complexity; this trade-off is governed by a
parameter that is optimized using cross-validation.

While cross-validation has many advantages for certain tasks, an often
mentioned disadvantage is that it is computationally expensive.
Indeed, $n$-fold cross-validation is typically implemented by running
the same learning system $n$ times, each time on a different training set
of size $(n-1)/n$ times the size of the original data set.
Because of this computational cost, cross-validation is sometimes avoided, 
even when it is agreed that the method would be useful.

It is clear, however, that when (for instance) a specific decision tree
induction algorithm is run several times on highly similar datasets, 
there will be redundancy in the computations.  E.g., when selecting 
the best test in a node of a tree, the test needs to be evaluated against
each individual example in the training set.  In an $n$-fold
cross-validation each example occurs $n-1$ times as a training
example, which means that each test will be evaluated on each
training example $n-1$ times.  The question naturally arises whether
it would be possible to avoid such redundant computations, thereby speeding
up the cross-validation process.  In this text we provide an affirmative
answer to this question.

This paper is organised as follows.  In Section 2 we focus on refinement
of a single node of the tree; we identify the computations that are
prone to the kind
of redundancy mentioned above, indicate how this redundancy can
be reduced, and analyse to what extent performance can thus be improved.
In Section 3 we discuss the whole tree
induction process, showing how our adapted node refinement algorithm
fits in several tree induction algorithms.  In Section 4 we present 
experimental results for one of these algorithms that support our 
complexity analysis, supporting our main claim that cross-validation can
be integrated with decision tree induction in such a way that it causes
only a small overhead. In Section 5 we briefly discuss to
what extent the results generalize to other machine learning techniques, 
and mention the limitations of our approach.  In Section 6 we conclude.

\section{Efficient Cross-validation}

\subsection{Decision Tree Induction}

\label{sec21}

We describe decision tree induction algorithms only in such detail
as needed for the remainder of this text, for more details see
Quinlan (1993) \nocite{Quinlan93:other} or Breiman et al. (1984).
\nocite{Breiman84:other}

\begin{figure}
\begin{tabbing}
xxx \= xxx \= xxx \= xxx \= xxxxxx \= \kill
{\bf function} {\sc grow\_tree}($T$: set of examples)\\
\> \> \> \> \>  {\bf returns} decision tree:\\
\> $t^*$ := optimal\_test($T$)\\
\> $\cal P$ := partition induced on $T$ by $t^*$\\
\> {\bf if} stop\_criterion($\cal P$)\\
\> {\bf then} {\bf return} {\bf leaf}(info($T$))\\
\> {\bf else} \\
\> \> {\bf for all} $P_j$ {\bf in} $\cal P$:\\
\> \> \>  $tr_j$ := {\sc grow\_tree}($P_j$)\\
\> \> {\bf return} {\bf node}($t^*$, $\bigcup_j \{(j, tr_j)\}$)\\
\end{tabbing}
\caption{\label{tdidt:alg}A generic algorithm for top-down induction of decision trees.}
\end{figure}

Decision trees are usually built top-down, using an algorithm similar
to the one shown in Figure~\ref{tdidt:alg}.  
Basically, given a data set, a node is created and a test is selected for
that node.  A test is a function from the example space to some finite
domain (e.g., the value of a discrete attribute, or the boolean
result of a comparison between an attribute and some constant).
Each test induces a partition of the data set (with each test result 
one subset is associated), and typically that test is selected
for which the subsets of the partition are
maximally homogeneous w.r.t. some target attribute (the ``class'', for 
classification trees).  For each subset of the partition, the procedure is
repeated and the created nodes become children of the current node.
The procedure stops when {\tt stop\_criterion} succeeds: this is typically
the case when no good test can be found or when the data set is sufficiently
homogeneous already.  In that case the subset becomes a leaf of the tree
and in this leaf information about the subset is stored (e.g., the 
majority class).  The result of the initial call of the algorithm is the
full decision tree.

The refinement of a single node (selecting the test and partitioning the 
data) can in more detail be described as follows:

\begin{tabbing}
xxx \= xxx \= xxx \= xxx \= \kill
{\bf for all} tests $t$ that can be put in the node:\\
\>  {\bf for all} examples $e$ in the training set $T$:\\
\> \> update\_statistics(S[$t$], $t(e)$, target($e$))\\
\> Q[t] := compute\_quality(S[t])\\
$t^*$ := argmax$_t$ $Q[t]$\\
partition $T$ according to $t^*$
\end{tabbing}

The computation of the quality of a test $t$ is split into two phases
here: one phase where statistics on $t$ are computed and stored into
an array $S[t]$, and a second phase where the quality is computed from
the statistics (without looking back at the data set).  For instance,
for classification trees, phase one could compute the class
distribution for each outcome of the test.\footnote{$S[t]$ is then a matrix
  indexed on classes and results of $t$, and
  update\_statistics($S[t]$, $t(e)$, $class(e)$) just increments
  $S[t]_{t(e),class(e)}$ by 1.}  Quality criteria such as information
gain or gain ratio \cite{Quinlan93:other} can easily be computed from
this in phase two.  For regression, using variance as a quality
criterion \cite{Breiman84:other}, a similar two-phase process can be
defined : the variance can be computed from $\sum (y_i^2, y_i, 1)$
where the $y_i$ are the target values.

\subsection{Removing Redundancy}

\subsubsection{Overlapping Data Sets}

Now assume that the node refinement process, as described above, is
repeated several times, each time on a slightly different data set $T_i$
(i.e., the $T_i$ have many examples in common).  We assume here
that the same set of tests is considered in all these nodes.  Then instead of
running the process $n$ times, with $n$ the number of data sets,
the following algorithm can be used:

\begin{tabbing}
xxx \= xxx \= xxx \= xxx \= \kill
{\bf for each} test $t$ that can be put in the node\\
\>  {\bf for each} example $e$ in $\bigcup_i T_i$: \\
\> \> {\bf for each} $i$ such that $e \in T_i$: \\
\> \> \> update\_statistics(S[$T_i$, $t$], $t(e)$, target($e$))\\
\> {\bf for each} $T_i$:\\
\> \> $Q[T_i, t]$ := compute\_quality($S[T_i,t]$)\\
{\bf for each} $T_i$: \\
\> $t^*_i$ := argmax$_t$ $Q[T_i,t]$\\
{\bf for each} {\em different} test $t^*$ among the $t_i^*$:\\
\> partition $\bigcup_i \{T_i | t^*_i = t^*\}$ according to $t^*$\\
\end{tabbing}

This algorithm performs the same computations as running the original one
once on each data set, except for two differences:
\begin{itemize}
\item for each test $t$, each single example $e$ is tested only once 
instead of $m(e)$ times, where $m(e)$ is the number of data sets the 
example occurs in.  
\item each single example $e$ is sorted into a child node\footnote{
Sorting examples into child nodes corresponds to partitioning the
data set.} $f(e)$ times, 
instead of $m(e)$ times, with $f(e)$ the number of different
best tests for all the data sets where the example occurred (obviously
$\forall e : f(e) \leq m(e)$).
\end{itemize}

Note that in each node of the tree multiple tests (at most $n$), and
correspondingly multiple sets of child nodes, may now be stored instead of
just one.

\subsubsection{Cross-validation}

For an $n$-fold cross-validation, each single example occurs exactly
$n-1$ times as a training example.  Hence, the time needed to compute
the statistics of all tests is reduced by a factor $n-1$ compared to 
running the original algorithm $n$ times.  The time needed to sort
examples into child nodes is reduced by $n-1$ if the same test is selected
in all folds, otherwise a smaller reduction occurs.  Besides this speedup
there are no changes in the computational complexity of the algorithm
(except for the extra computations involved in, e.g., selecting elements
from a two-dimensional array instead of a one-dimensional array).

Specifically for cross-validation, the algorithm can be further improved
if the employed statistics $S$, for any data set $D$, can be computed from
the corresponding statistics of its subsets in a partition.  This holds
for all statistics that are essentially sums (such as those mentioned
in Section~\ref{sec21}), since in that case $S(D) = \sum_i S(D_i)$.  
Such statistics could also be called {\em additive}.

In an $n$-fold cross-validation, the data set $D$ is partitioned into
$n$ sets $D_i$, and the training sets $T_i$ can be defined as $D -
D_i$.  It is then sufficient to compute statistics just for the
$D_i$; those for the $T_i$ can be easily computed from this without further
reference to the data (first compute $S(D) = \sum_i S(D_i)$; then
$S(T_i) = S(D) - S(D_i)$).  Since each example occurs in exactly 1 of the
$D_i$, updating statistics has to be done only $N$ times instead
of $N(n-1)$ times (with $N$ the number of examples).

\subsubsection{Cross-validation Combined with Actual Tree Induction}

In practice, cross-validation is usually performed in addition to
building a tree from the whole data set: this tree is then considered
to be the actual hypothesis proposed by the algorithm, and the
cross-validation is done just to estimate the predictive accuracy of
the hypothesis or for parameter tuning.  The algorithm for efficient
cross-validation can easily be extended so that it builds a tree from
the whole data set in addition to the cross-validation trees (just add
a virtual fold 0 where the whole data set is used as training set; note
that $S(T_0) = S(D)$).
Adopting this change, we obtain the algorithm in
Figure~\ref{finalalg}.  In the remainder of this text we will refer to
this algorithm as the {\em parallel algorithm}, as opposed to the
straightforward method of running all cross-validation folds and the
actual tree induction serially (the {\em serial algorithm\/}).

\begin{figure}
\begin{tabbing}
xxx \= xxx \= xxx \= xxx \= \kill
\{ $D$ is the set of all examples relevant for this node,\\
\> partitioned into $n$ subsets $D_i$, $i=1 .. n$.  \\
\> $T_0 = D$, and for $i>0$ $T_i = D - D_i$ \}\\
1. \> {\bf for each} test $t$ that can be put in the node\\
2. \> \>  {\bf for each} example $e$ in $D$: \\
3. \> \> \> choose $i$ such that $e \in D_i$ \\
4. \> \> \> update\_statistics($S[D_i, t]$, $t(e)$, target($e$))\\
5. \> \> compute $S[T_i, t]$ ($i=0..n$) from all $S[D_j, t]$\\
6. \> \> {\bf for each} $T_i$:\\
7. \> \> \> $Q[T_i, t]$ := compute\_quality($S[T_i,t]$)\\
8. \> {\bf for each} $T_i$ :\\
9. \> \> $t^*_i$ := argmax$_t$ $Q[T_i,t]$\\
10. \> {\bf for each} {\em different} test $t^*$ among the $t_i^*$:\\
11. \> \> partition $\bigcup_i \{T_i | t^*_i = t^*\}$ according to $t^*$\\
\end{tabbing}
\caption{\label{finalalg}Performing cross-validation in parallel with 
induction of the actual tree.}
\end{figure}



At this point, we have discussed the major issues related to the refinement
of a single node.  The next step is to include this process into a full
tree induction algorithm.  This will be discussed in the next section,
but first we take a look at the complexity of the node refinement step.

\subsection{Computational Complexity of Node Refinement}

Let $t_e$ be the time for extracting relevant information from a single 
example (i.e., the example's target value and test result)
and updating the statistics matrix $S$ (in other words, executing line
4 in the algorithm in Figure~\ref{finalalg} once);
$t_p$ the time needed to test an example and sort it into the correct
subset during partitioning;
$N$ the number of examples in the data set, $n$ the 
number of folds, and $a$ the number of tests.  Then we obtain the following
times for refining a single node (the $c_i$ denote terms constant in $N$):
\begin{itemize}
\item when building one tree from the full data set:
$T_{\rm actual} = a N t_e + N t_p + c_1  = N ( a t_e + t_p) + c_1$
\item when performing cross-validation serially:\\
$T_{\rm 1\ fold} = {n-1 \over n} N (a t_e + t_p) + c_2$\\
$T_{\it n\ \rm folds} = (n-1) N (a t_e + t_p) + c_3$
\item when serially building the actual tree and performing a 
cross-validation:\\
$T_{\rm serial} = T_{\rm actual} + T_{n\ \rm folds} = n N (a t_e + t_p) + c_4$
\item when using the parallel algorithm, worst case (all folds select 
different tests):\\
$T_{\rm parallel} = a N t_e + n N t_p + c_5 = N(a t_e + n t_p) + c_5$
\item when using the parallel algorithm, best case (all folds select
the same test):\\
$T'_{\rm parallel} = N (a t_e + t_p) + c_6$
\end{itemize}

Our analysis gives rise to approximate upper bounds on the speedup factors
that can be achieved.  Assuming large $N$ so that the $c_i$ terms can be
ignored (hence ``approximate''), for the worst case we get
\[
{T_{\rm serial} \over T_{\rm parallel}} = n {a t_e + t_p \over a t_e + n t_p} < n
\]
and
\[ 
{T_{\rm serial} \over T_{\rm parallel}} = {a t_e + t_p \over {a t_e \over n} + t_p}
< {a t_e + t_p \over t_p} = 1 + a {t_e \over t_p}\]

Hence the worst case speedup factor is bounded by $\min(n, 1+a
t_e/t_p)$.  It will approximate $n$ when a) $N$ becomes large and b)
$t_p$ is small compared to $a t_e$.  In the best case, where the same
test is selected for all folds, we just get 
$T_{\rm serial} / T'_{\rm parallel} < n$: the speedup factor approaches $n$
as soon as $N$ becomes large.  Another way to look at this is
to observe that
$T'_{\rm parallel} / T_{\rm actual}$ approaches one; in other words,
for large $N$ and a stable problem (where small perturbations in
the data do not lead to different tests being selected) the overhead
caused by performing cross-validation becomes negligible.

\section{An Algorithm for Building Trees in Parallel}

We now describe how the above algorithms for node refinement fit in 
decision tree induction algorithms.  First we describe the data structures,
which are more complicated than when growing individual trees.
Next we discuss several decision tree induction techniques and 
show how they can
exploit the above algorithms.

\subsection{Data Structures}

Since the parallel cross-validation algorithm builds multiple trees at
the same time, we need a data structure to store all these trees
together.  We refer to this structure as a ``forest'', although this
might be somewhat misleading as the trees are not disjoint, but may share
some parts.  

\begin{figure}
\epsfig{file=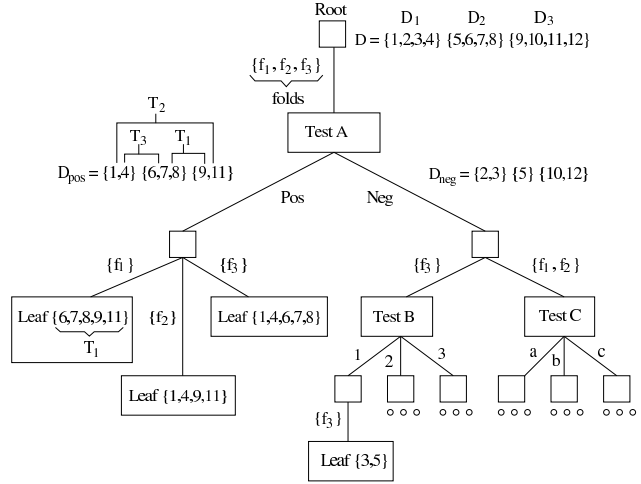,width=8.5cm}
\caption{\label{forest:fig}An example forest for a 3-fold cross-validation.}
\end{figure}

An example of a forest is shown in Figure~\ref{forest:fig}.
In this figure two kinds of internal nodes are represented.
The small squares represent bifurcation points, points where the trees of
different folds start to differ because different tests were selected.
The larger rectangles represent tests that partition the relevant data set.
The way in which the trees in the forest split the data sets is illustrated
by means of an example data set of 12 elements on which a three-fold 
cross-validation is performed.

Note that the memory consumption of a forest is (roughly) at most $n+1$ 
times that of a single tree (this happens when at the root different tests
are obtained for all $n$ folds plus the actual tree), which in practice is
not problematic since $n$ usually is small.

When in the following we refer to nodes in the forest,
we always refer to the test nodes, making abstraction of bifurcation points.
E.g., in Figure~\ref{forest:fig} the root node has five children,
three of which are leaves.

\subsection{Tree Induction Algorithms}

\subsubsection{Depth First Tree Induction}

Probably the best known approach to decision tree induction is
Quinlan's (1986) \nocite{Quinlan86:jrnl} ID3 algorithm, later developed
into C4.5 \cite{Quinlan93:other}.  ID3 basically follows the depth-first 
approach of Figure~\ref{tdidt:alg}.

The simplest way to adapt an ID3-like algorithm to perform cross-validation
in parallel with the actual tree building, is to make it use the node 
refinement algorithm of Figure~\ref{finalalg} and call the algorithm 
recursively for each child node created.  Note that the number of such
child nodes is now $\sum_{i=1}^f r_i$, with $f$ the number of different
tests selected as best test in some fold and $r_i$ the number of possible
results of the $i$-th test.


In this way, the above mentioned speedup is obtained as long as the same test
is chosen in all cross-validations and in the actual tree.  The more different
tests are selected, the less speedup is achieved; and when in each fold
a different test is selected, the speedup factor goes to 1 (all folds are
handled separately).  

To see how this process influences the total forest induction time,
let us define $t_r(i)$ as the average time that is needed to refine all the
nodes of a single tree on level $i$ for a data set of size $|D|$, 
and $f(i)$ as the average number of different tests selected on level $i$
of the forest (averaged over all nodes on that level of the forest).  The 
computational complexity of the whole forest building process can then be
approximated as
\[ T_{\rm parallel} = t_r(1) + f(1) t_r(2) + f(2) t_r(3) + \cdots \]
for the parallel version, and, assuming 
that refinement time is linear in the number of examples in nodes
that are to be refined,\footnote{From this it follows that in one
fold of $n$-fold cross-validation the actual refinement time for level $i$ is 
${n-1 \over n} t_r(i)$.}
\[ T_{\rm serial} = n t_r(1) + n t_r(2) + n t_r(3) + \cdots \]
for the serial version (we obtain $n t_r(i)$ and not $(n+1) t_r(i)$ because
the $n$ folds have size ${n-1 \over n} |D|$).

Thus the total speedup will be between 1 and $n$, and will be higher
for stable problems (low $f(i)$) than for unstable problems (most $f(i)$ 
close to $n+1$).

\subsubsection{Level-wise Tree Induction}

Most decision tree induction algorithms assume that all data reside in
main memory.  When inducing a tree from a large database, this may not
be realistic: data have to be loaded from disk into main memory when
needed, and then for efficiency reasons it is important to minimize
the number of times each example needs to be loaded (i.e., minimize
disk access).  To that aim alternative tree induction algorithms have
been proposed \cite{Mehta96:proc,Shafer96:proc} that build the tree one
whole level at a time, where for each level one pass through the data
is required.  The idea is to go over the data and for each
example, update statistics for all possible tests in the node (of the 
currently lowest level of the tree) where the example belongs.  
For each node the best test is then selected from these statistics 
without more access to the data.

Since in these approaches, too, the computation of the quality of tests
is split up into two phases (computing statistics from data, computing 
test quality from statistics), it is easy to see how such level-wise
algorithms can be adapted.  When processing one example, instead of looking
up the single node in the tree where the example belongs, one should look up
all the nodes in the forest where the example belongs (for an example not
yet in a leaf this is at least one node and at most $n-1$ nodes, with $n$ 
the number of folds) and update the statistics in all these nodes.

When data reside on disk, the number of examples is typically large
and both $t_e$ and $t_p$ are large (due to external data access).  The
constant terms $c_i$ then become negligible very quickly, 
and the speedup factor can approach $n$ if $a \geq n {t_p \over t_e}$.  
Assuming that $t_p$ and $t_e$ are comparable, this will be true as soon
as $a \geq n$, which in practice often holds.





\subsection{Further optimisations}

As soon as different tests are selected for different folds, the forest
induction process bifurcates in the sense that from that point onwards
different trees in the forest will be handled independently.
A further optimisation that comes to mind, is removing redundancy among
computations in these independently handled trees as well.

Referring to Figure~\ref{forest:fig}, among the different branches
created by a bifurcation point (square node) there may still be some
overlap with respect to the tests that will be considered in the child
nodes, as well as the relevant examples.  For instance, in the lower
right of the forest in Figure~\ref{forest:fig}, in the children of the
``test B'' node one needs to consider all tests except A and B, and in
the children of the test C node one needs to consider all tests except
A and C.  Since the relevant example set for fold $f_3$ at that point
(\{2,3,5\}) overlaps with that of folds $f_1$ and $f_2$
(\{2,3,5,10,12\}), all tests besides $A$, $B$ and $C$ will give rise
to some redundant computations.

Removing this redundancy as well would give rise to a more thorough
redesign of the forest induction process; it seems that for best
results the depth-first tree induction method should be
abandoned, and a level-wise method adopted
instead.  Here we will not discuss this optimisation any further but
focus on the above described algorithm, which is simple and compatible
with both tree induction approaches and can easily be integrated in
existing tree induction systems.

\section{Experimental Evaluation}

\subsection{Implementation}

We implemented Algorithm \ref{finalalg} as a module of
{\sc Tilde} \cite{Blockeel98b:jrnl}, an ILP system (inductive logic programming \cite{Muggleton94-JLP:jrnl}) that induces
first order decision trees; briefly, these are decision trees where a test
in a node is a first order literal or conjunction, and 
a path from root to leaf can be interpreted as a Horn clause.
Literals belonging to different nodes in such a path may share variables.

A typical property of ILP systems in general, and {\sc Tilde} is no
exception, is that because tests are first order conjunctions, both the 
number of tests and the time needed to perform a test 
may be large.  This translates to large $a$, $t_e$ and $t_p$ values in
our complexity analysis, which makes it reasonable to expect a speedup
factor close to $n$ for refinement of the top node of the tree; and
close to $n/f(i-1)$ for nodes on level $i$.

\subsection{Experimental Setup}

For these experiments we used the version of {\sc Tilde} as implemented
within the ACE data mining tool\footnote{ACE is available
for academic purposes upon request.} \cite{Blockeel00:proc}; this version
is a depth-first ID3-like algorithm that keeps all data in main memory.

With these experiments we aim at a better understanding of the behaviour
of the parallel cross-validation process.  We measure how much speedup the
parallel procedure yields, compared to the serial one; how the overhead
of the parallel procedure varies with the number of folds; and how much time
is spent by both procedures on different levels of the tree.


The parallel and serial procedures make use of exactly the same
implementation of {\sc Tilde} except for the differences between 
parallel and serial execution as described in this text.
The different procedures are compared pairwise for the following data
sets:
\begin{itemize}
\item {\bf SB} (Simple Bongard) and {\bf CB} (Complex Bongard): several
  artificially generated sets of so-called ``Bongard'' problems 
  \cite{Deraedt95-ALT:proc} (pictures are classified according to simple
  geometric patterns).  SB contains 1453 examples with a simple
  underlying theory, CB 1521 examples with a more complex theory.
\item {\bf Muta}: the Mutagenesis data set \cite{Srinivasan96:jrnl}, an ILP benchmark
(230 examples)
\item {\bf ASM}: a subset of 999 examples of the so-called ``Adaptive Systems
Management'' data set, kindly provided to us by Perot Systems Nederland.
\item {\bf Mach}: ``Machines'', a tiny data set (15 examples) described in
\cite{Blockeel98b:jrnl}
\end{itemize}
The number of tests in each node varied from 3 to a few hundred (as tests
are first-order clauses, their number may vary greatly even among nodes of 
the same tree).



\subsection{Results}

\begin{figure}
\epsfig{file=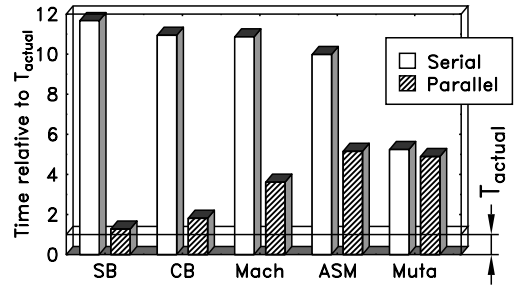}
\caption{\label{timings:fig}$T_s$ and $T_p$ relative to $T_a$.  The part above
the horizontal line is the overhead $O_s$ respectively $O_p$.}
\end{figure}

Table~\ref{timings:tab} compares the actual tree building time $T_a$,
the time for serially performing 10-fold cross-validation in addition
to the actual tree building $T_s$, and the time needed by the parallel
algorithm $T_p$.  In addition to these, the speedup factor $S = T_s / T_p$ 
is shown as as well as the overhead caused by performing the
cross-validation ($O_s = 100 (T_s / T_a - 1) \%$, similarly for $O_p$).  
$O_s$ and $O_p$ are plotted graphically in Figure~\ref{timings:fig}.

\begin{table}
\begin{tabular}{|l|llllll|} \hline
      & $T_a$   & $T_s$   & $T_p$     & $S$    & $O_s$    & $O_p$ \\ \hline
SB  & 2.6       & 31      & 3.4       & 9.2      & 1100\%   & 27\%       \\
CB  & 4.0                     & 44            & 7.4             & 6.0
            & 1000\%        & 81\%       \\
Mach & 0.028                  & 0.30          & 0.10            & 3.0  
            & 990\%         & 260\%      \\
ASM & 720                     & 7100          & 3700            & 1.9
            & 900\%         & 420\%      \\
Muta & 1300                   & 6700          & 6200            & 1.1
            & 420\%         & 390\%      \\ \hline
\end{tabular}
\caption{\label{timings:tab}Timings of parallel and serial execution on 
various data sets (in seconds).}
\end{table}

The lowest overhead is achieved for Simple
Bongard, which has a relatively large number of examples and a simple
theory.  The simplicity of the true theory causes the induced trees to
be exactly the same in most folds, yielding little bifurcation.  For
Complex Bongard, the effect of bifurcation is more prominent.  For
ASM, a real-world data set for which a perfect theory may not exist,
the overhead of cross-validation is relatively high (but still better
than for the serial algorithm).  For Machines, the overhead is relatively
large but still smaller than for the serial algorithm; i.e., even for 
small example sets the parallel algorithm yields a speedup.

\begin{figure}
\epsfig{file=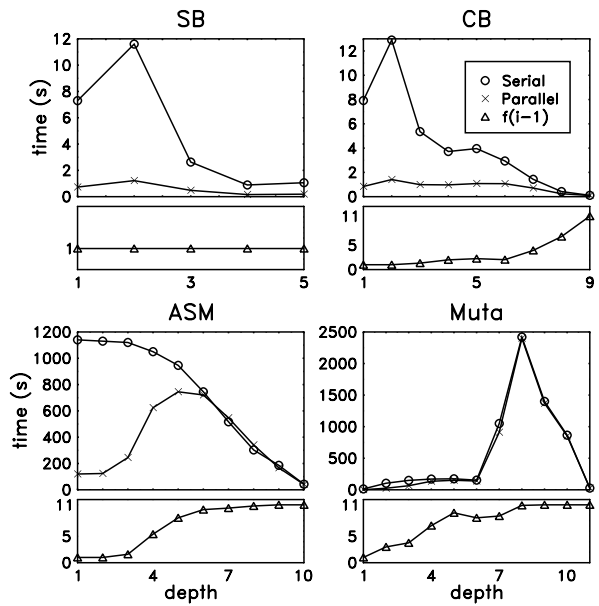}
\caption{\label{nfold-fi:fig}Total refinement time per level.}
\end{figure}

For Mutagenesis we obtained less good results.  Two factors turned out
to be responsible for this: instability of the trees, but also high
variance in the complexity of testing examples.  The latter is due to the
fact that first-order queries have exponential worst-case complexity;
most of them are reasonably fast, but a very few of them may dominate
the others, time-wise.  Such behaviour typically occurs at lower
levels of the tree, as will be confirmed when we look at
Figure~\ref{nfold-fi:fig}.

\begin{figure}
\epsfig{file=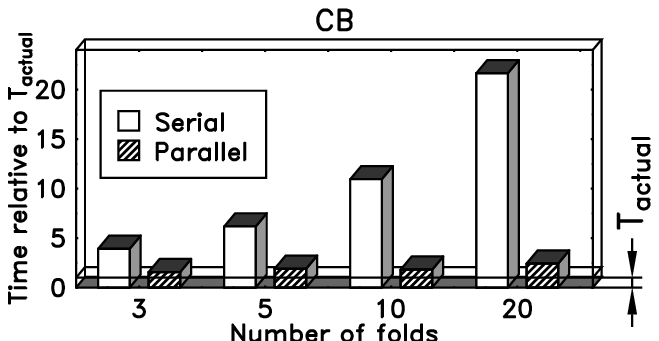}
\epsfig{file=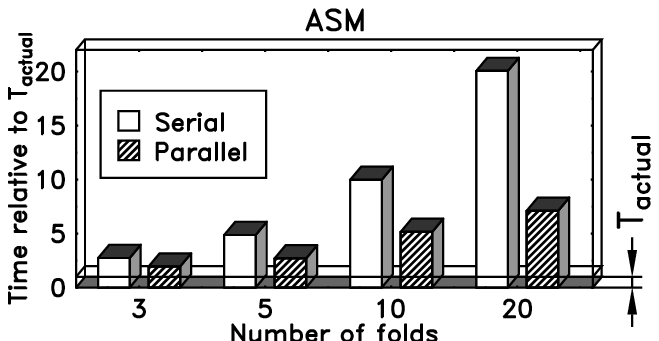}
\caption{\label{nfolds:fig}Overhead in function of number of folds}
\end{figure}

Figure~\ref{nfolds:fig} shows how cross-validation overhead varies
with the number of folds for the CB and ASM data sets.
The result for CB confirms our expectation that $n$ has a small
influence on the total time, but for ASM the overhead increases with 
increasing $n$.  

The latter result can be understood by looking at the graphs in
Figure~\ref{nfold-fi:fig}, where the total time spent on each level of
the tree by the parallel and the serial procedure is plotted, together
with the $f(i-1)$ values as defined previously.  The graphs clearly
show that when $f$ goes up, the per-level speedup factor is reduced.
For CB, this happens at a point where the total refinement
time is already small, so it does not influence the overall
speedup factor much; but for ASM and Muta $f$ increases almost
immediately.  Note that in the part where $f$ is
high, many folds are handled independently and cross-validation
becomes linear in $n$, which explains the increase of the ASM data in 
Figure~\ref{nfolds:fig}.  It is also clear in Figure~\ref{nfold-fi:fig} 
how the time spent on some lower levels suddenly goes up; this is the
effect of stumbling upon some very complex tests.



\section{Applicability and Limitations}

Although we have studied efficient cross-validation in the context of
decision trees, the principles explained here are also applicable outside
this domain.  For instance, rule set induction systems typically build a rule
by consecutively adding a ``best'' condition to it until no further 
improvement occurs.  Similar to our forest-building algorithm,
cross-validation of such rules could be performed in parallel with the 
construction of the actual rule set, avoiding redundant computations.

It is less clear, however, how the technique could be used with models
that contain only continuous parameters, such as neural networks.  We
obtain the greatest speedups for stable trees, where the same test is
chosen in different folds.  With continuous models, no computations
will ever be exactly the same, hence removal of exactly redundant
computations as explained here will in general not be possible.

Also within decision tree induction a number of limitations exist.
A first one is related to the use of continuous parameters in the tree.
Decision tree induction systems often construct inequality
tests for continuous attributes (e.g., $A<5.3$) where the constant is 
generated from the currently relevant data.  Even for stable problems 
where the same test is usually selected for different folds, there may 
be small differences in the constants that make the tests look different.  
Solving this problem requires extra optimisations.

A second limitation is that the proposed techniques concern the tree
building phase only.  This phase is typically followed by tree
post-pruning, and may be preceded by data pre-processing, such as
discretization of attributes \cite{Fayyad93:proc}.  While these other
phases usually take much less time than the tree building phase, when
they are not negligible and $n$ is large they may become the
bottleneck, limiting the usefulness of our approach (unless
optimisations similar to the ones discussed here are also possible in
these phases).

\section{Conclusions}

We have shown that in the context of decision tree induction the
benefits of cross-validation are available for a relatively low
overhead, if the cross-validation is carefully integrated with the
normal tree building process.  Comparing experimental results with
an analytical estimate of this overhead, we have identified a number
of disturbing factors, such as variance in test complexity (which causes
variance in the overhead) and tree instability (which causes the overhead to 
increase on average).  These factors increase the overhead, but in all
cases it was still smaller than for the serial cross-validation procedure,
and in the best cases there was only a small overhead over the normal
tree induction process.


The ideas underlying our approach are also applicable outside the decision
tree context, e.g., for rule induction, but not immediately
for induction of models that have only continuous parameters.

Possible further improvements to the technique include specific adaptations
for handling tests with continuous values.  Also, the algorithms we have
discussed are fairly simple versions; the SPRINT system for instance 
\cite{Shafer96:proc} is much more sophisticated with respect to the
statistics it keeps, and adaptations to the system along the lines of this
paper would be correspondingly complex to implement.

Related work includes that of Moore and Lee (1994)
\nocite{Moore94:proc}, who discuss efficient cross-validation in the
context of model selection.  Their approach differs substantially from
ours in that they obtain efficiency by quickly abandoning models that
after seeing some examples have low probability of ever becoming the
best model; i.e., they save on the number of cases a model is
evaluated on during cross-validation, whereas our work focuses on
removing redundancy in the model building process itself.

Blockeel et al.\ (2000) \nocite{Blockeel00:proc} discuss a technique
similar to the one described here.  The main difference is is in the
kind of redundancies that are removed; here the redundancies arise
from running the same test in different folds of a cross-validation,
whereas in Blockeel et al.\ (2000) \nocite{Blockeel00:proc} they are
caused by similarities in different tests (the tests being first-order
conjunctions, which might be similar up to one literal).  Both approaches
can easily be combined, and such work is in progress.

\section*{Acknowledgements} 
 
The authors are a post-doctoral fellow, respectively research assistant,
of the Fund for Scientific Research of Flanders (Belgium).  
They thank Perot Systems Nederland / Syllogic for providing the
ASM data.  The cooperation between Perot Systems Nederland and
the authors was supported by the European Union's Esprit Project 28623 (Aladin).



\end{document}